\documentclass{article}

\usepackage[final, nonatbib]{neurips_2024}

\usepackage[utf8]{inputenc} 
\usepackage[T1]{fontenc}    
\usepackage{url}            
\usepackage{booktabs}       
\usepackage{amsfonts}       
\usepackage{nicefrac}       
\usepackage{microtype}      
\usepackage{xcolor}         
\usepackage{multirow}
\usepackage{amsmath}
\usepackage{graphicx}
\usepackage[colorlinks=true, allcolors=blue]{hyperref}
\usepackage{threeparttable}
\usepackage{wrapfig}
\usepackage{tikz}
\usepackage{subcaption}
\usepackage[export]{adjustbox}
\usepackage{graphbox}

\title{Open-Source  Acceleration of Stable-Diffusion.cpp Deployable on All Devices}

\author{Jingxu Ng$^1$, Cheng Lv$^{2,4}$,  Pu Zhao$^2$, Wei Niu$^3$,  Juyi Lin$^2$, \\ \textbf{Minzhou Pan$^2$, Yun Liang$^1$,  Yanzhi Wang$^{2,4}$} \thanks{Corresponding author} 
\\
\\$^1$Peking University, $^2$Northeastern University, $^{3}$University of Georgia, $^{4}$SealAI}

\begin{document}
\maketitle

\begin{abstract}
Stable diffusion plays a crucial role in generating high-quality images. However, image generation is time-consuming and memory-intensive. To address this, stable-diffusion.cpp (Sdcpp) emerges as an efficient inference framework to accelerate the diffusion models. Although it is lightweight, the current implementation of ggml\_conv\_2d operator in Sdcpp is suboptimal, exhibiting both high inference latency and massive memory usage. To address this,  
in this work, we present an optimized version of Sdcpp leveraging the Winograd algorithm to accelerate 2D convolution operations, which is the primary bottleneck in the pipeline. By analyzing both dependent and independent computation graphs, we exploit the device's locality and parallelism to achieve substantial performance improvements. Our framework delivers correct end-to-end results across various stable diffusion models, including SDv1.4, v1.5, v2.1, SDXL, and SDXL-Turbo. Our evaluation results demonstrate a speedup  up to  \textbf{2.76$\times$}  for individual convolutional layers and an inference  speedup up to \textbf{4.79$\times$} for the overall image generation process, compared with the original Sdcpp, on the same M1 pro.
\\
Homepage: \textit{https://github.com/SealAILab/stable-diffusion-cpp}
\end{abstract}

\section{Introduction}

Stable diffusion \cite{rombach2022high} has established itself as a powerful tool for generating high-quality images. However, the computational demands of image generation pose significant challenges, particularly in terms of inference latency and memory consumption. To address these limitations, stable-diffusion.cpp (Sdcpp) \cite{leejet2024stable-diffusion-cpp} emerges as an efficient inference framework to accelerate the diffusion models.   Sdcpp is a C/C++ implementation of the stable diffusion model, designed for efficient inference on CPUs (and potentially GPUs with the appropriate configuration) without external dependencies. The implementation is based on GGML \cite{ggerganov-ggml}, which works in the same way as llama.cpp \cite{ggerganov-llama.cpp}.  It is lightweight  without external dependencies. However, the current implementation of the computation-intensive 2D convolution operator in Sdcpp remains inefficient, incurring massive inference latency. To mitigate this problem, we apply the Winograd algorithm~\cite{lavin2015fastalgorithmsconvolutionalneural} to optimize convolution operations in Sdcpp  for faster inference and  reduced  memory cost. Specifically, we analyze both dependent and independent computation graphs, leveraging the device’s locality and parallelism to achieve substantial performance gains in convolution operations.  Our evaluations across various models and image sizes demonstrate our significant improvements in terms of the inference speed. Notably, we can achieve an inference acceleration  up to  \textbf{2.76$\times$}  for individual convolution layers and a  speedup up to \textbf{4.79$\times$}  for the whole image generation process, compared with the original Sdcpp on M1 pro. The homepage is \textit{https://github.com/SealAILab/stable-diffusion-cpp}.

\section{Techniques}

As the current implementation of 2D convolution operator in Sdcpp is relatively slow with high memory usage,  we apply  the Winograd algorithm to optimize convolution operations in Sdcpp, speeding up the generation process and reducing the computation and memory costs.
With Winograd, the  convolution operation is split into multiple steps: (i) preprocess of filter and activation weights, (ii) element-wise multiplication between the preprocessed  tensor, and (iii) postprocess of the intermediate results. In this work, we analyze both the dependent and independent computation graphs, leveraging the device’s locality and parallelism to achieve substantial performance gains in convolution operations.

To enhance locality, we apply scatter-store and gather-load optimizations, ensuring that the data placement fits within the L1 cache during the loading process, thereby minimizing cache swapping. For parallelism, we exploit the independent operations within the Winograd algorithm, distributing independent computations across multiple threads and cores to reduce image generation latency. In fact, GGML~\cite{ggerganov-ggml} employs a shared-state approach, which assigns cores with the same workload. Instead, in our optimization, upon finishing their current computations, the computation cores dynamically receive the next computational block. This action can benefit the calculation on computational units with P-core (performance core) and E-core (efficiency core), such as the M-series Macs. By dynamically assigning workloads based on core performance, the approach ensures efficient utilization and balanced computation across various computation cores.

\section{Evaluation}

\textbf{Operator Support.}  Previously, a number of operators in the original Sdcpp are not supported in Android, leading to incorrect image generation results for certain SD models such as SDXL. 
Currently, in our optimization, we have supported all operators in multiple SD models for various devices such as Mac, Android, and AMD devices. We can also support the operators for the diffusion transformer models, which are widely used in video generation models such as Open-Sora \cite{opensora}. 
Besides, the quantization of the operators is also supported in our implementation. In future work, we will further optimize the speed of these operators in the common operator set.

\textbf{Model Support.} We have implemented an end-to-end acceleration pipeline of stable diffusion based on our operator library. Our framework can generate correct end-to-end results for all SDv1.4, v1.5, v2.1, SDXL, and SDXL-Turbo, tested on Mac (GPU, Metal~\cite{metal})  and Android (Qualcomm, OpenCL~\cite{5457293}) devices. 
It can also support other variants such as Realistic Vision and user-specified arbitrary LoRA~\cite{hu2022lora} modules as well. 

\subsection{Single Convolution Layer Speedup Performance}

We demonstrate the performance of our convolution operator implementation using the Winograd algorithm for multiple convolutional layers from the Sdcpp sampling process as shown in Table~\ref{tab:single}. Specifically, we focus on the layers which are adopted more frequently in the sampling process. As demonstrated in Table \ref{tab:single}, our optimization can lead to the inference speedup of above $2\times$ for various convolutional layers under different configurations.

\begin{center}
\begin{threeparttable}[t]
\caption{Latency comparison  with Sdcpp for various convolution layers.}\label{tab:single}
\begin{tabular}{c|c}
\hline Convolution Layer \\Filter = [KW, KH, IC, OC], Activation = [IW, IH, IC, N]                                           & Our speedup \\ 
\hline Filter = [3, 3, 640, 640] \\ Activation = [32, 32, 640, 1]           & 2.76x     \\
\hline Filter = [3, 3, 1280, 1280] \\ Activation = [16, 16, 1280, 1]        & 2.27x     \\
\hline Filter = [3, 3, 2560, 1280] \\ Activation = [16, 16, 2560, 1]        & 2.20x     \\
\hline Filter = [3, 3, 320, 320] \\ Activation = [64, 64, 320, 1]           & 2.09x     \\
\hline Filter = [3, 3, 640, 320] \\ Activation = [64, 64, 640, 1]           & 2.07x     \\
\hline
\end{tabular}
\end{threeparttable}
\end{center}

\subsection{Overall Speedup Performance for Image Generation}
We demonstrate the inference speedup performance for image generation on M1 pro and M2 max in Table \ref{tab:M1} and Table \ref{tab:M2} respectively. As observed, our method achieves faster inference speed compared with Sdcpp under different configurations of various image sizes and models. Specifically, when the image size becomes larger such as $1024\times 1024$, our improvements over Sdcpp is more significant (such as the $4.79\times$ speedup for FP32 in M1 pro), demonstrating our superior acceleration performance.

\begin{center}
\begin{threeparttable}[t]
\caption{Latency comparison  with Sdcpp for various models on M1 pro with 16GB memory and MacOS Sonoma 15.1.}
\label{tab:M1}
\begin{tabular}{c|c|c|c|c}
\hline
Model                       & Steps               & Image size & Type  & Our speedup \\ \hline
\multirow{2}{*}{SDXL}       & \multirow{2}{*}{20} & \multirow{2}{*}{1024$\times$1024}                           & F32  & 4.79$\times$   \\ \cline{4-5} 
                            &                     &                                                      & F16   & 3.06$\times$    \\ \hline
\multirow{2}{*}{SDv2}        & \multirow{2}{*}{20} & \multirow{2}{*}{768$\times$768}                             & F32  & 2.02$\times$    \\ \cline{4-5} 
                            &                     &                                                      & F16  &  1.68$\times$   \\  \hline
\multirow{2}{*}{SDv1.5}      & \multirow{2}{*}{20} & \multirow{2}{*}{512$\times$512}                             & F32    & 1.84$\times$   \\ \cline{4-5} 
                            &                     &                                                      & F16    & 1.51$\times$    \\\hline
\end{tabular}
\end{threeparttable}
\end{center}

The speedup performance on M2 max is demonstrated in Table~\ref{tab:M2}.  We can make similar observations that  our method achieves significant  inference speedup compared with Sdcpp under various image sizes for different models. Similarly, with the image size of $1024\times 1024$, our speedup over Sdcpp for the SDXL model can be as large as  $4.64\times$  for FP32 in M2 max, demonstrating our superior acceleration performance. 

\begin{center}
\begin{threeparttable}[t]
\caption{Latency comparison  with Sdcpp for various models on M2 max with 32GB memory and MacOS Sequoia 15.0.}
\label{tab:M2}
\begin{tabular}{c|c|c|c|c}
\hline
Model                       & Steps               & Image size & Type  & Our speedup \\ \hline
\multirow{2}{*}{SDXL}       & \multirow{2}{*}{20} & \multirow{2}{*}{1024$\times$1024}                    & F32      & 4.64$\times$   \\ \cline{4-5} 
                            &                     &                                                      & F16      & 3.12$\times$    \\ \hline
\multirow{2}{*}{SDv2}        & \multirow{2}{*}{20} & \multirow{2}{*}{768$\times$768}                      & F32      & 1.95$\times$    \\ \cline{4-5} 
                            &                     &                                                      & F16      & 1.73$\times$   \\  \hline
\multirow{2}{*}{SDv1.5}      & \multirow{2}{*}{20} & \multirow{2}{*}{512$\times$512}                      & F32      & 1.77$\times$   \\ \cline{4-5} 
                            &                     &                                                      & F16      & 1.53$\times$    \\\hline
\end{tabular}
\end{threeparttable}
\end{center}

\subsection{Visualization}

\begin{figure}[t]
 \centering
\begin{tabular}{p{0.1in}p{0.85in}p{0.85in}p{0.85in}p{0.85in}p{0.85in}}

& 
\parbox{0.85in}{\centering \footnotesize a lovely cat} &  
\parbox{0.85in}{\centering \footnotesize a man playing a guitar}  &  
\parbox{0.85in}{\centering \footnotesize mountains, river, and trees}  & 
\parbox{0.85in}{\centering \footnotesize futuristic city}&
\parbox{0.85in}{\centering \footnotesize an apple and a banana}

\\

\vspace{-0.4in} \rotatebox{90}{\centering \footnotesize Origin sdcpp }  &
\includegraphics[align=c,width=1in]{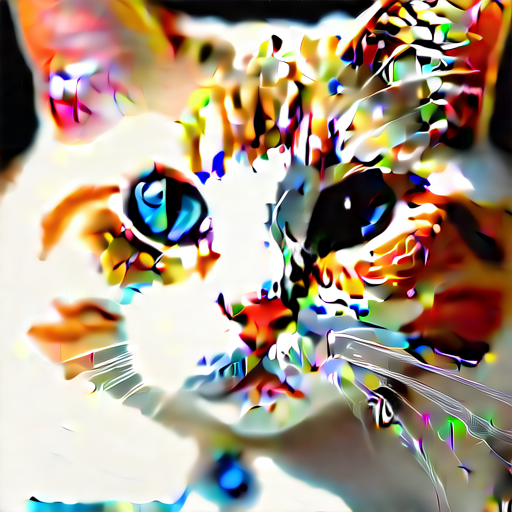} &
\includegraphics[align=c,width=1in]{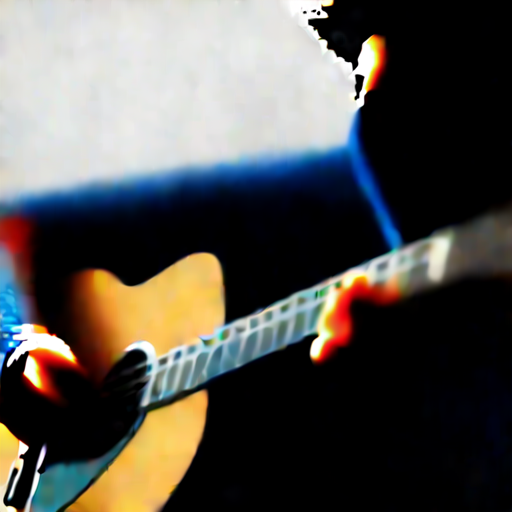} & 
\includegraphics[align=c,width=1in]{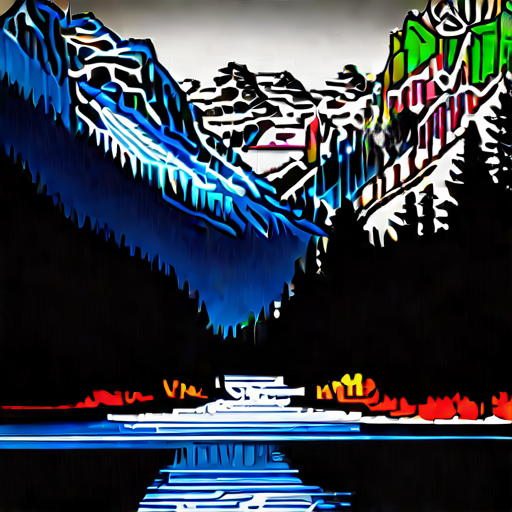} &
\includegraphics[align=c,width=1in]{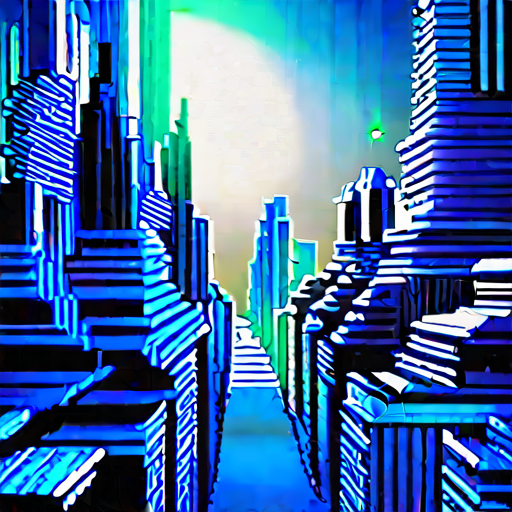} &
\includegraphics[align=c,width=1in]{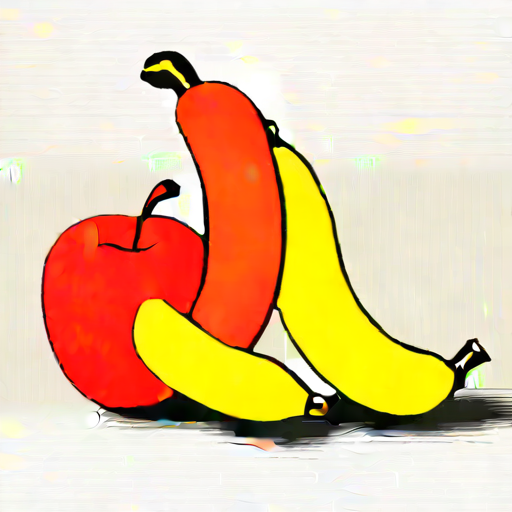}

\\ 

\vspace{-0.4in} \rotatebox{90}{\parbox{0.6in}{\centering \footnotesize Ours}} & 

\includegraphics[align=c,width=1in]{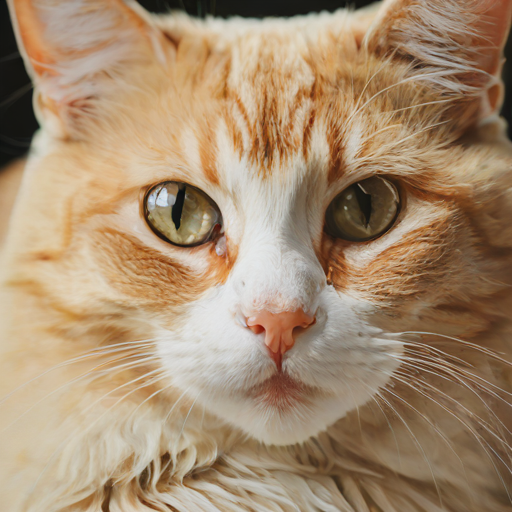} &
\includegraphics[align=c,width=1in]{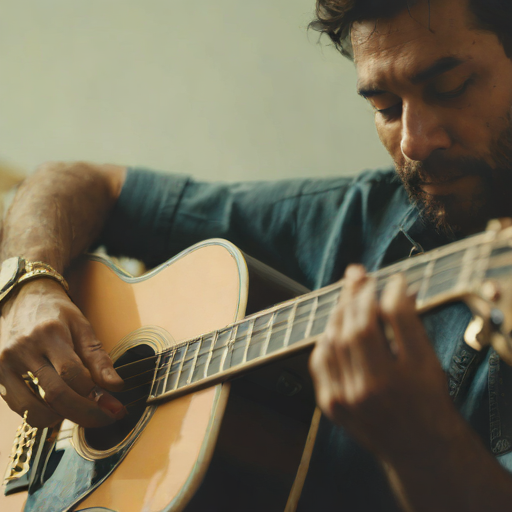} & 
\includegraphics[align=c,width=1in]{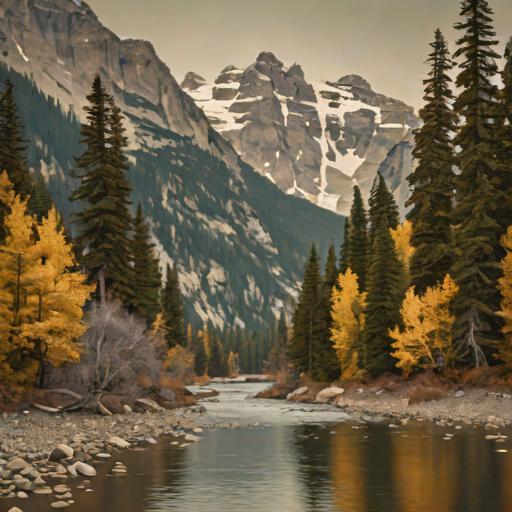} &
\includegraphics[align=c,width=1in]{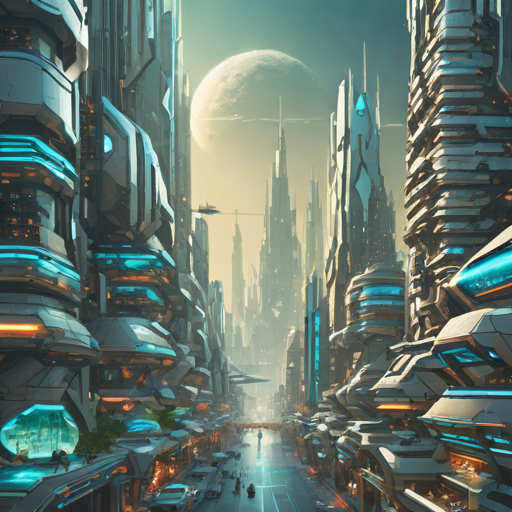} &
\includegraphics[align=c,width=1in]{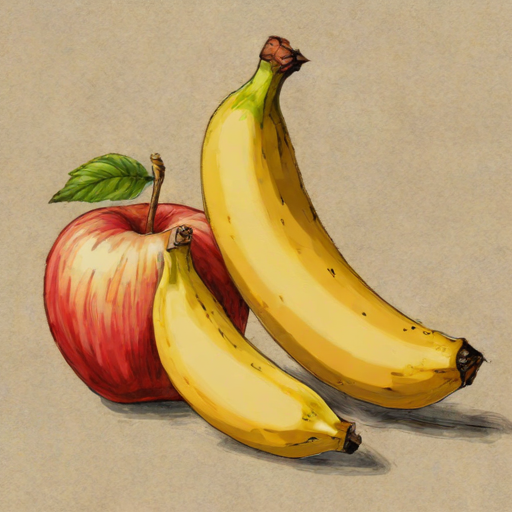}

\end{tabular}
\caption{Visualization examples of the original Sdcpp and ours, with SDXL-Turbo model and 5 steps.} 
\label{fig:examples}
\end{figure}

We demonstrate some generation examples from the original Sdcpp and ours in Figure \ref{fig:examples}, with the SDXL-Turbo model and 5 steps for both. We can observe that with our improvements, the generated images are more realistic than those of the original Sdcpp, under the same prompt.

\section{Conclusion}
The current implementation of ggml\_conv\_2d in Sdcpp remains slow and memory-intensive. We optimize Sdcpp using the Winograd algorithm to overcome these limitations. Our enhanced framework supports end-to-end image generation for all tested models, including SDv1.4, v1.5, v2.1, SDXL, and SDXL-Turbo, consistently delivering correct results. Comprehensive testing across these models highlights the substantial speedup achieved by our optimizations.

\bibliographystyle{unsrt}
\bibliography{sample}

\end{document}